\documentclass[conference]{IEEEtran}
\IEEEoverridecommandlockouts
\usepackage{amsmath,amssymb,amsfonts}
\usepackage{algorithmic}
\usepackage{graphicx}
\usepackage{textcomp}
\usepackage{xcolor}
\usepackage[utf8]{inputenc} 
\usepackage[T1]{fontenc}    
\usepackage{hyperref}       
\usepackage{url}            
\usepackage{booktabs}       
\usepackage{amsfonts}       
\usepackage{nicefrac}       
\usepackage{microtype}      
\usepackage{adjustbox}
\usepackage{biblatex}
\def\BibTeX{{\rm B\kern-.05em{\sc i\kern-.025em b}\kern-.08em
    T\kern-.1667em\lower.7ex\hbox{E}\kern-.125emX}}
\begin{document}

\title{Input-length-shortening and text generation via attention values}

\author{\IEEEauthorblockN{1\textsuperscript{st} Neşet Özkan TAN}
\IEEEauthorblockA{\textit{School of Computer Science} \\
\textit{School of Computer Science}\\
School of Computer Science \\
neset.tan@auckland.ac.nz}
\and
\IEEEauthorblockN{2\textsuperscript{nd} Alex Yuxuan Peng}
\IEEEauthorblockA{\textit{School of Computer Science} \\
\textit{The University of Auckland}\\
Auckland, New Zealand \\
ypen260@aucklanduni.ac.nz}
\and
\IEEEauthorblockN{3\textsuperscript{rd} Joshua Bensemann}
\IEEEauthorblockA{\textit{School of Computer Science} \\
\textit{The University of Auckland}\\
Auckland, New Zealand \\
josh.bensemann@auckland.ac.nz}
\and
\IEEEauthorblockN{4\textsuperscript{th} Qiming Bao}
\IEEEauthorblockA{\textit{School of Computer Science} \\
\textit{The University of Auckland}\\
Auckland, New Zealand \\
 qbao775@aucklanduni.ac.nz}
\and
\IEEEauthorblockN{5\textsuperscript{th} Tim Hartill}
\IEEEauthorblockA{\textit{School of Computer Science} \\
\textit{The University of Auckland}\\
Auckland, New Zealand \\
thar011@aucklanduni.ac.nz}
\and
\IEEEauthorblockN{6\textsuperscript{th} Mark Gahegan}
\IEEEauthorblockA{\textit{School of Computer Science} \\
\textit{The University of Auckland}\\
Auckland, New Zealand \\
m.gahegan@auckland.ac.nz}

\and
\IEEEauthorblockN{7\textsuperscript{th} Michael Witbrock}
\IEEEauthorblockA{\textit{School of Computer Science} \\
\textit{The University of Auckland}\\
Auckland, New Zealand \\
m.witbrock@auckland.ac.nz}
}
\maketitle

\begin{abstract}
 Identifying words that impact a task's performance more than others is a challenge in natural language processing. Transformers models have recently addressed this issue by incorporating an attention mechanism that assigns greater attention (i.e., relevance) scores to some words than others. Because of the attention mechanism's high computational cost, transformer models usually have an input-length limitation caused by hardware constraints. This limitation applies to many transformers, including the well-known bidirectional encoder representations of the transformer (BERT) model. In this paper, we examined BERT's attention assignment mechanism, focusing on two questions: (1) How can attention be employed to reduce input length? (2) How can attention be used as a control mechanism for conditional text generation?We investigated these questions in the context of a text classification task. We discovered that BERT's early layers assign more critical attention scores for text classification tasks compared to later layers. We demonstrated that the first layer's attention sums could be used to filter tokens in a given sequence, considerably decreasing the input length while maintaining good test accuracy. We also applied filtering, which uses a compute-efficient semantic similarities algorithm, and discovered that retaining approximately 6\% of the original sequence is sufficient to obtain 86.5\% accuracy. Finally, we showed that we could generate data in a stable manner and indistinguishable from  the original one by only using a small percentage (10\%) of the tokens with high attention scores according to BERT's first layer. 
\end{abstract}


\section{Introduction}
In recent years, transformer-based pre-trained language models (PLM), also known as foundation models \cite{found}, have achieved state-of-the-art results on a variety of tasks in the field of Natural Language Processing (NLP). PLMs are often trained on a large corpus of data, such as Wikipedia articles, news, and books, to capture the context of the corpus in a self-supervised manner. They require significant hardware resources to optimise the model's parameters \cite{few}. In this process, an input (a set of words) is pre-processed into tokens (words, sub-words, or characters), each token corresponding to a multi-dimensional vector representation. Like other parameters in the model, the vector representations of tokens change with respect to a loss function during the training process and are stable during inference for downstream tasks.

BERT, or Bidirectional Encoder Representations from Transformers is one such PLM, which has achieved high results in recent years  \cite{bert}. BERT is an example of the transformer architecture \cite{atten}, which uses transformer blocks. The key novelty of the Transformer block is the use of the attention mechanism \cite{atten}, where self-attention ``heads'' assign a relevance score for every token in the sequence with respect to the rest of the tokens via attention calculations.
These calculations work by projecting token vectors onto $d$-dimensional key $\mathbf{K}$, query $\mathbf{Q}$, and value $\mathbf{V}$ vectors, then taking the following dot products of these projections for each head.
\begin{equation}\label{atq}
 { \operatorname{Attention}( \mathbf{Q}, \mathbf{K}, \mathbf{V} )
 =
 \operatorname{softmax}\left(\frac{\mathbf{Q K}^{\top}}{\sqrt{d}}\right) \mathbf{V}}
\end{equation}
The calculated attention scores are used for BERT's two training objectives: (1) Predicting the next sentence and (2) predicting a masked word. The masked language modelling helps to learn an internal representation of the vector representations by masking 15\% of the given sequence, and the bidirectional structure considers each token in the context of the entire sequence instead of the words appearing before it \cite{bert}.
 
The number of following models that are direct descendants of BERT demonstrates its significance in the field. Examples of these descendants include XLNet \cite{xl}, {RoBERTa} \cite{roberta}, ALBERT \cite{Albert}, SciBERT \cite{scibert} and BioBERT \cite{biobert}. RoBERTa is a replication of BERT that explores the impact of several critical hyperparameters and the training data amount. ALBERT was developed using strategies to reduce the number of parameters of BERT so that it could run faster with less accuracy loss. XLNet is an extended pretraining method that maximises the learning abilities of bidirectional contexts and overcomes the constraints of BERT due to its original training formulation. There are also domain-adapted versions of BERT, which are trained on specific domains such as SciBERT and BioBERT. SciBERT, in particular, was trained on a vast corpus of scientific articles from several scientific fields. The application area for these versions of BERT includes, the protein folding problem \cite{prot}, image classification \cite{img}, and generative networks \cite{gen}.

Equation \ref{atq} is the main contributor to PLMs computational complexity since it has quadratic complexity and repeatedly occurs in the transformer-based model's architecture. Due to this high computational cost, transformer-based models usually limit the maximum length of the input sequence (typically 512 tokens). Designing transformer-based architectures that allow longer inputs has recently become an active and competitive research area \cite{inf}, \cite{long}, \cite{bigbert}, and \cite{etc}. The main aim of these studies is to reduce time and memory costs by modifying the self-attention mechanism. However, here we have taken a different approach which leads us to our first research question: How can the attention scores of tokens be used to shorten the input in a text-classification task?

We investigated two methods to select the words/sub-words in a sequence to shorten input length. More precisely, we applied two filtration methods to the IMDB dataset \cite{imdb}: (1) Filtering based on BERT's first layer's attention scores. (2) Similarity-based filtering is used by eliminating the most similar sentences in a sequence. Then, we fine-tuned the version of BERT in \cite{hug} according to these new filtered datasets. Even though the new training set consisted of filtered tokens that were less than half the length of the full-length sequences, we achieved close accuracy proximity to the full-length trained model in both cases. In the first case, the accuracy was only around 2\% lower than full-sequence, while in the second case, the accuracy was 1\% lower than the full-length fine-tuning regime. We also tested the shortening idea in a specific domain(scientific papers) for multi-class classification tasks and obtained a similar result to that in the binary classification task. We will discuss these outcomes in Section \ref{filt}.

Our second research question is based on another well-known PLM, the second version of Generative Pre-trained Transformer (GPT-2)  \cite{GPT2}. GPT-2, like BERT, was trained on millions of sentences taken from the internet, and it performs remarkably well in reading comprehension, translation, and summary tasks \cite{GPT2}. Unlike BERT's bidirectional objective, GPT-2 calculates attention by considering only the words that  come before the given word in a phrase.

We investigated the following question by utilising GPT-2's generative power. Can attention scores of BERT be used as a control mechanism for text generation via GPT-2?
We used pre-trained GPT-2 for imitating data points conditioned on a certain proportion of the tokens with the highest attention scores according to BERT's first layer. In other words, we fine-tuned GPT-2 for text generation under the control of BERT's first-layer attention scores. Then we generated data points under different input designs,  such as by only inputting [tokens] and  [label + tokens]. We imitated the reviews with the desired label indistinguishable from the original one. The processes and results of the conditional text generation will be detailed in Section \ref{imt}.

\section{The Use of Attention for Filtering}\label{filt}
\subsection{Attention score-based filtering}
    In this section, we aimed to determine whether it is possible to reduce the length of a sequence without significantly sacrificing model performance. We initially considered BERT's attention scores in layers to answer this question. We used the IMDB dataset for the sentiment prediction task. We measured the accuracy of train and test datasets via a pre-trained BERT model, which was fine-tuned with full-length text.
   \begin{figure}[htbp] \label{allay}
   \vskip 0.05in
   \centering
\includegraphics[scale=0.4]{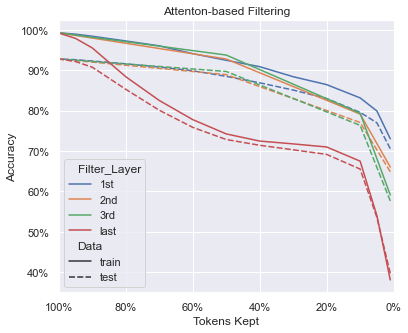}
\caption{The insight for accuracy of filtered sequences regarding attention assignments of BERT's initial layers versus the final layers. The assignments of the early layers result in better accuracy than later ones.} 
\vskip 0.2in
\end{figure} 
During the filtering progress,  the layers of the BERT encoder were used to extract the attention weights generated from each sample. A single attention matrix was created by adding the cumulative attention weights from the 12 attention heads.  The sum of each of the matrix's columns was then calculated, creating an attention score for each token. There are two special tokens in each sequence, namely CLS and SEP tokens to indicate start and end of sequence. The CLS and SEP tokens were removed, and the top X percentile of tokens was selected. The CLS and SEP tokens were appended at the start and end of the new sequence, respectively, and the sequence was input into the model to predict its sentiment.

We discovered that tokens chosen by BERT's initial layers are more effective than tokens chosen by later layers.We executed all of the filtration operations using BERT's first layer because it is the best option in terms of low computing cost for filtering, Figure 1. (The 12-layer version of the figure is included in the Appendix).

Using pre-trained BERT (without fine-tuning fpr the sentiment prediction task), we selected the top-50\% and bottom-50\% tokens of each sequence and then fine-tuned pre-trained BERT with these filtered datasets. Finally, we compared the models' accuracy by testing full-length sequences (Figure 2). We observed that fine-tuning with tokens with higher attention scores improves the fine-tuned model's accuracy compared to the tokens with low attention scores.

\begin{figure}[htbp]\label{50}
\centering
\includegraphics[scale=0.4]{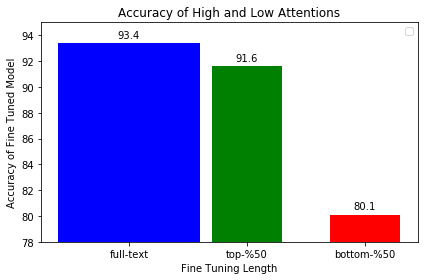}
\caption{The full-length test data accuracy for the fine-tuned model with full-length, top-$50\%$ and bottom-$50\%$, respectively. The width of the rectangles demonstrates the lengths of the sequences during the fine-tuning process}. 

\end{figure}

\subsection{Similarity-based filtering}
To see whether there are other efficient filtering methods rather than using the direct attention  BERT model, we eliminated the most similar sentences for each sequence by using their sentence embedding. We used Sentence-BERT (SBERT)  \cite{SBERT} to find semantic textual similarities between sentences. SBERT is a pre-trained BERT network that uses siamese and triplet network architectures to generate semantically relevant sentence embeddings to score and rank sentence pairs \cite{SBERT}. This way, we obtained semantic textual similarities lower cost than the original BERT embedding, which has quadratic complexity for similarity calculations. A comparison of BERT and  SBERT was conducted by  \cite{SBERT}, and similarity computations were dramatically reduced, from 65 hours to 5 seconds.
In our experiment, the longer sentences were eliminated, and only short sentences were kept for each sequence. We were able to eliminate 53\% of each sequence this way. With this dataset, we fine-tuned BERT, which consisted of 47\% of the length of the original sequences. On the full-length test set, we obtained 92.6\% accuracy. We compared the same filtration rates by considering BERT's first layer selection. Then, we repeated the same process up to 
 $6\%$ reduction rate. The comparison between similarity-based filtering with BERT-base filtering is shown in Figure 3.

\begin{figure}[htbp]\label{SBERT}
\begin{center}
\includegraphics[scale=0.40]{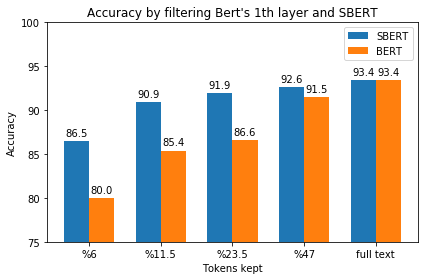}
\caption{Similarity-based filtering(blue) and BERT's first layer-based filtering.} 
\end{center}
\end{figure} 
\subsection{Input-shortening for various domains}
To further investigate the generalizability of our findings, we examined the impact of reducing sequence length across various domains and more complex tasks beyond binary sentiment classification in the general domain (IMDB reviews). In addition, we constructed a scientific paper dataset for multi-class classification tasks by using abstracts from scholarly papers in four distinct fields: computer science, mathematics, biology, and physics. We extracted and cleaned the abstracts of articles from the arxiv dataset introduced in \cite{arxiv} to create the dataset, selecting 40,000 data points based on the "category" feature, which assigns a distinct sub-field name to each paper by the authors.

Using 30,000 abstracts and label pairs obtained from the above progress, we fine-tuned the SciBERT\footnote{The model's checkpoints were taken from the HuggingFace repository introduced in \cite{hug}.} model to complete  multi-class classification tasks. Simultaneously, we shortened abstracts by using the SciBERT model's attention scores, and then we fine-tuned the SciBERT model with the 50\% shorter data points. The final models were tested on 10,000 full-length data points, and their accuracy was compared in Table~\ref{scibert}. We obtained less than 1\% accuracy loss by cutting the abstract length in half.

We also applied the shortening method for the verdict prediction task, which is a label prediction task for claims based on evidence provided (such as supported or refuted). For this, we used the well-known fact checking dataset FEVER \cite{Thorne2018-nx}. We only considered statements that were supported or refuted, and we limited the sample size of the dataset to obtain balanced classes. We end up with around 60K claim and evidence pairs, half of which were labelled as supported and the other half as refuted. We used the BERT model's attention scores to shorten the concatenation of claim and evidence pairs, and then we fine-tuned the BERT model with the 50\% shorter data points and  compared it to the full-length fine tuning regime. The final models were tested on about 20K full-length data points, and the accuracy for full length and half-length was 92\% and 90\%, respectively.

\begin{table}
\centering
\begin{tabular}{l|r}
{\normalsize Sequence length} & {\normalsize  Accuracy} \\\hline
{\normalsize  Full-length} & {\normalsize  92.03}\\
{\normalsize  Half-length} & {\normalsize  91.34}\\
\end{tabular}
\caption{\label{scibert} The full-length test data accuracy for the fine-tuned SciBERT model compared with  top-$50\%.$}
\end{table}

\section{The Use of Attention for Text Generation}\label{imt}
This section delves into whether the attention scores of tokens extracted from the first layer of BERT can be utilized for text generation.
\subsection{Text generation}
We combined BERT's attention scores with the generative power of GPT-2. In other words, we used BERT's attention for conditional text generation. To the best of our knowledge, this is the first research that uses attention scores as a control mechanism for text generation in a multi-pre-trained language model setting. We retained the top $10\%$ and $20\%$ tokens from each IMDB sample (there are $50, 000$ samples in this dataset) according to the first layer of BERT. We used  $40,000$ samples to fine-tune GPT2 on conditional text generation. The design of training was the following.
\begin{center}
\textbf{[sentiment +randomized-top-tokens (obtained by BERT) + target (full text)]}

\end{center}
We utilized the fine-tuned GPT2 model to generate 10,000 reviews based on the top 10\% and 20\% of tokens. We explored two input styles: (1) [sentiment + randomized-top-tokens] following the fine-tuning regime and (2) [randomized-top-tokens] with the sentiment component left empty. Subsequently, we assessed the accuracy of the fine-tuned BERT model for sentiment analysis on the generated data examples. Remarkably, we achieved nearly the same accuracy as the original data points for the first input type, as presented in Table~\ref{acc_generated}.

\begin{table}
\begin{tabular}{l|r}
Text & BERT's Accuracy \\\hline
Gold (original) text & 93.4 \\
Imitated-reviews from top-$20\%$ tokens & 93.07\\
Imitated-reviews from top-$10\%$ tokens  & 93.04\\
Imitated-reviews from top-$20\%$ tokens without sent & 75.4\\
Imitated-reviews with top-$10\%$ tokens without sent & 73.1
\end{tabular}
\caption{\label{tab:widgets} The accuracy measured by fine-tuned BERT model for original test data and imitated test data sets, which are conditioned on top-tokens and sentiments.}
\label{acc_generated}
\end{table}
\subsection{Evaluation }
We evaluated the resulting text's cohesion and fluency. We randomly sampled 100 data points (50  generated and 50 from the original dataset). Two evaluators, both were  native English speakers, evaluated each generated text without knowing whether it was the synthetic or original text. The evaluators were requested to assign a score between 0 and 5 for cohesiveness, taking into account the following two cohesion principles that were given in \cite{will}.\\

\noindent\fbox{%
    \parbox{\columnwidth}{%
    {\small
\textbf{Principle 1}: A cohesive paragraph has consistent topic strings.

\textbf{Principle 2}: A reader will feel that a paragraph is cohesive if it has other strings of related words, which we will call thematic strings.
    }}%
}\\

In addition, the evaluators were asked to give the text a fluency score between 0 and 5 based on how well-formed the English text appeared to them. After this, the average of the allocated ratings was calculated. The average fluency and cohesiveness of the original text were $2.66$ and $3.42$, respectively. On the other hand, the mean fluency and cohesiveness of generated text were $2.92$ and $3.48$, respectively. In other words, the evaluators score the cohesiveness and fluency of the generated text slightly higher than the original.

We also evaluated the generated text in a scalable and automated manner. We calculated a BERTScore (proposed by \cite{BERTscore}) for each entry. The BERTScore algorithm calculates a similarity score for each pair of candidate and reference phrases by considering contextual embeddings (BERT embeddings) rather than precise matches. More precisely, according to   \cite{BERTscore} for each token $x_i$ in a reference text $x$, the following precision, recall and  $F_1$  scores are calculated by considering tokens $\hat{x}_i$ in a generated text $\hat{x}$:
$${\displaystyle  R_{BERT}=\frac{1}{|x|}\sum_{x_i\in x}\max_{\hat{x}_j\in x}x_i^T\hat{x}_j,}$$  $${\displaystyle P_{BERT}=\frac{1}{|\hat{x}|}\sum_{\hat{x}_j\in \hat{x}}\max_{x_i \in x}x_i^T\hat{x}_j,}$$ $${\displaystyle F_{BERT}=2\frac{P_{BERT}R_{BERT}}{P_{BERT}+R_{BERT}}.}$$
By following the above formulations, we calculated precision, recall, and $F_1$ scores for each pair of original and generated texts by considering the same sample given to human evaluators. Then we calculated the mean of each metric score, which was obtained from each pair. The precision was 0.77, the recall was 0.79, and the $F_1$ was 0.78. One example of generated and synthetic texts is shown below.

\subsection*{Example}
The first paragraph that follows is an example generated using the top tokens of the second paragraph. The GPT-2 model was fine-tuned with the specified input design and was used to generate the first text. The bold words in the first text are in the top ten percent of the original text's tokens (the second text).\\

\noindent\fbox{%
    \parbox{\columnwidth}{%
        {\small I have \textbf{occasionally} seen \textbf{Jerry} Lewis in some of his more \textbf{amusing} \textbf{films}, but this is one of the funniest \textbf{comedies} I have ever seen. The premise is \textbf{somewhat} similar to THE KID IN THE RAIN ( \textbf{yes}, it's about a little girl and her \textbf{enemy} ) - except that there is no child at all in \textbf{the} \textbf{film}. It's an \textbf{amusing} \textbf{comedy} from beginning to end, and even has a couple of gags that are not nearly as \textbf{funny} as they should be. In fact, most of \textbf{the} humor comes from Lewis' predictable mannerisms, which make him seem like a caricature of himself.  He doesn't need to do anything really special to make this \textbf{movie} work; \textbf{you} just have to enjoy seeing him play so well.}
    }%
}
\bigskip

\noindent\fbox{%
    \parbox{\columnwidth}{%
        {\small Written by brilliant Monkees'TV writers Gerald Gardner and Dee Caruso, WHICH WAY TO THE FRONT was the last of the " Jerry Lewis " movies until " Hardly Working " almost a decade later. Jerry's comedy is evidently an acquired taste, and admittedly he can occasionally be his own worst enemy when he helms as producer director -  but even in the dreariest of his films, there are always moments of brilliance.  WHICH WAY manages to be amusing, entertaining and yes, quite funny. It is somewhat unlike any of the typical Lewis films. The pace is very upbeat and the are lots of excellent supporting players  a kind of JERRY DOES HOGANS HEROES. The whole thing looks kind of like an unsold TV pilot and you will either love it or hate it but hopefully YOU WILL LAUGH.}
    }%
}

\bigskip
\section{Related Work}
There has been substantial recent research on examining the attention mechanism. Layer-based attention distribution analysis for 128-token-long inputs was conducted in \cite{man} to measure the syntactic ability of attention heads. One of the findings of \cite{man} is that the self-attention heads within the same layer have the similar attention distribution. A similar result was obtained in \cite{Mic}, where they argued that a reasonable amount of attention heads could be removed during test time without significant performance loss. According to \cite{low}, BERT's initial layers are crucial for capturing word-order information. In contrast, middle layers are essential for syntactic information \cite{man2} and the final layer representations are prominent for task-specific adaptation \cite{pr}. However, the relationship between attention weights and model outputs is ambiguous. For example, \cite{nop} finds that the attention values have weak correlation with feature importance measures using gradient or feature erasure methods. They also demonstrate that different sets of attention values learned using adversarial training can result in the same prediction, therefore attention values should not be utilised as an explanation of the model's predictions. Although attention values cannot be considered as the ``exclusive'' explanation for the model's predictions, \cite{wiegreffe2019attention} argues that attention values are still ``plausible'' explanation of the model's predictions. They also show that the alternative attention values obtained through adversarial training do not perform as well when used in a diagnostic MLP model. It is important to note that both \cite{nop} and \cite{wiegreffe2019attention} study the attention mechanism in RNN-based models, instead of Transformer-based large-scale pretrained language models such as BERT that was used in our experiments.

Token dropping has been investigated recently as an approach to improving the efficiency of Transformer-based models. For instance, \cite{hou2022token} specifically explores token dropping during pretraining BERT. They report that their method reduces the pretraining cost by 25\% without significant suffering in performance on downstream tasks. Our work differs in that we use the attention scores obtained from pretrained BERT to decide which token to drop during the fine-tuning stage. Both \cite{guan2022transkimmer} and \cite{guan2022block} investigate token dropping across the hidden layers and on downstream tasks. However, they do not improve the efficiency of the fine-tuning process. They only perform ``skimming'' during inference on downstream tasks.

Generating long and informative reviews conditioned on contexts is challenging. Many approaches have been explored to tackle this problem. For example, in \cite{Rei}, a statistical algorithm was designed to generate sentiment phrases by considering the co-occurrence of words. The model named SentiGAN \cite{wang} applies Generative Adversarial Networks (GAN) to generate diverse texts using Monte Carlo search. Self-attentive recursive auto-encoders were used in \cite{li} to create a model that takes product information, user reviews, and their writing styles as input to generate controlled and individualised reviews. However, the computational complexity of all of the models above may be excessively high in the case of long text generation tasks, resulting in unsatisfactory results. Recently, Transformer-based language models have been applied to generating texts for sentiment analysis tasks. For example, \cite{der} uses T5~\cite{raf} to generate texts given pseudo sentences/phrases (similar to templates) that contain sentiment information. Prompt-tuning is another approach that makes use of pretrained language models to generate texts conditioned on contexts. For instance, \cite{lil} designs prompts that contain information on aspects, opinions, and polarities of sentiments, and use the prompts as contexts for text generation. To our knowledge, no study has investigated using attention weights to identify important tokens and use these tokens as contexts for conditional text generation. 
 
\section{Conclusion}
We investigated BERT's attention weights for two goals in this study: (1) shortening input length and thus saving training costs, and (2) generating new examples with  a desired sentiment. We used the attention weights and embeddings of BERT's first layer in our experiments because of the lower computational cost and the experimental results showing that the early layers are more useful for filtering tokens while maintaining good accuracy. We also evaluated a similarity-based filtering strategy at the sentence level for the first goal by removing longer sentences with similar semantics as the shorter ones. We achieved higher accuracy with this strategy than filtering tokens according to attention weights in BERT's first layer. The models trained on data with almost half of the tokens removed could achieve the similar test accuracy as the model trained on full-length data. A similar outcome was concluded for the verdict prediction task. We further investigate  input shortening for multi-class classification task on scientific paper corpus which shows that the attention scores of the first layer can be used for shortening input in the scientific domain and beyond binary text classification. Additionally, we demonstrated that we could generate high-quality new examples by using BERT's first layer to select a small proportion of the tokens with high attention scores. These examples, which are indistinguishable from the original one according to human evaluators and generated text, have a reasonably high precision, recall, and F1 scores according to the BERTscore-metrics.
\appendix
\subsection*{Experimental Setup}
In our BERT model fine-tuning experiments, we use the original BERT  and SciBERT checkpoints provided by HuggingFace \cite{hug}. These models include 12 layers and 12 transformer blocks in each layer, the hidden layer size is 768, and the pre-trained model has 110 million parameters. For the generation part, we used "gpt2-medium" 24 layers with 16 transformer blocks, the hidden layer size is 1024, and the pre-trained model has 345 million parameters. The  number of model parameters that we used in this work is shown in the table  \ref{par}.
\subsection*{Computing sources}
In all of our experiments, we used a single NVIDIA Quadro RTX 8000 graphics processing unit with 48GB of RAM capacity.

\begin{table}[htbp]
\centering
\begin{tabular}{l|l|}
    Models &  Parameters    \\\hline
    BERT &  109M    \\
    SciBERT&  110M\\
    GPT-2& 345M
\end{tabular}
\caption{Parameters per model}
\label{par}
\end{table}
\subsection*{Hyper-parameters}
During the training generation model, we used maximum 1024 for the max length of the sequence to be generated. We used a $5e-4$ learning rate with a $1e-8$ EPS and the warm-up step was $1e2. $ The epoch number for generation part was $5$ and we did experiment with  $16$ batches. During inference time, we generated text between $100$ and $520$ maximum length.   The number of highest probability vocabulary tokens to keep for top-k-filtering was 30 and we applied Top-p (nucleus) sampling at a rate $0.7$. The model temperature parameter was $0.9$ with a $3.0$ reputation penalty. Early stopping was inputted as "True" and we returned a single sequence.  

During fine-tuning BERT with IMDB data, we used the default parameters of the shared model in the Huggingface platform\footnote{https://huggingface.co/.}. More details can be found in the related page. \footnote{https://huggingface.co/bert-base-uncased.} In the IMDB experiment, we retrieved the dataset from the dataset library\footnote{https://huggingface.co/docs/datasets/index.} in the same platform.

We used a variant of the Adam optimizer (AdamW) with a 3e-5 learning rate and $0.01$ weight decay for training the arxiv dataset for multi-classs classification task. The validation split was  $0.3$ and we ran the experiment with $5$ epochs and  $16$ batches.
\subsection*{All layers attention}
  \begin{figure}[htbp] \label{allay}
   \vskip 0.05in
   \centering
\includegraphics[scale=0.4]{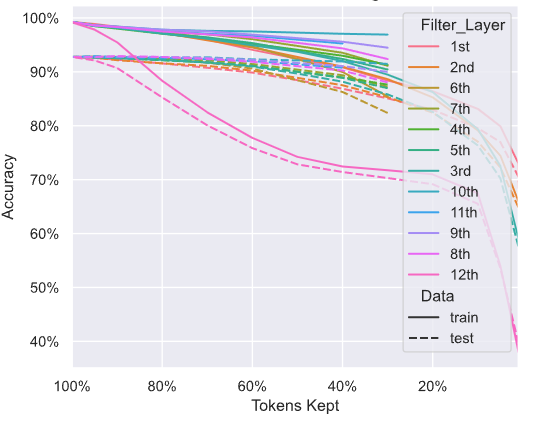}
\caption{The insight for accuracy of filtered sequences regarding attention assignments of BERT's initial layers versus the final layers. The assignments of the early layers result in better accuracy than later ones.} 
\vskip 0.2in
\end{figure}

\printbibliography
\end{document}